\documentclass{article}

\usepackage[nonatbib,preprint]{neurips_2020}
\usepackage{graphicx}
\usepackage[]{algorithm2e}

\usepackage{adjustbox}
\usepackage[utf8]{inputenc} % allow utf-8 input
\usepackage[T1]{fontenc}    % use 8-bit T1 fonts
\usepackage{hyperref}       % hyperlinks
\usepackage{url}            % simple URL typesetting
\usepackage{booktabs}       % professional-quality tables
\usepackage{amsfonts}       % blackboard math symbols
\usepackage{nicefrac}       % compact symbols for 1/2, etc.
\usepackage{microtype}      % microtypography
\usepackage{color}
\usepackage{xcolor}
\usepackage{ulem}
\usepackage{amsmath}
\usepackage{mathtools}
\usepackage{graphicx,subfigure}

\newcommand{\norm}[1]{\left\lVert#1\right\rVert}

\title{Cassandra: Detecting Trojaned Networks from Adversarial Perturbations}

\author{
    Xiaoyu Zhang\thanks{Center for Research in Computer Vision, University of Central Florida} \\
    \texttt{x.zhang@knights.ucf.edu} \\
    \And
    Ajmal Mian\thanks{Computer Science and Software Engineering, The University of Western Australia} \\
    \texttt{ajmal.mian@uwa.edu.au} \\
    \And
    Rohit Gupta\textsuperscript{*} \\
    \texttt{rohitg@knights.ucf.edu} \\
    \And
    Nazanin Rahnavard\thanks{Department of Electrical and Computer Engineering, University of Central Florida} \\
    \texttt{nazanin@eecs.ucf.edu} \\\
    \And
    Mubarak Shah\textsuperscript{*} \\
    \texttt{shah@crcv.ucf.edu} \\
}

\begin{document}

\maketitle

\begin{abstract}
%Deep neural network has been widely used as the main building block in many critical applications thanks to  its demonstrated strong capabilities in image classification and recognition. However, the potential security risks threaten large scale deployment and application of DNN models, some of critical problems are still unsolvable. For example, the vulnerabilities of neural networks. We proposed a two stage deep learning frame work to detect trojans in a pre-trained model based on the back propagated perturbations that, we believe, can capture the fingerprint of inserted backdoor. The first stage, we train a detector to classify clean and trojanned models. In the second stage, an anomaly detector was applied to discriminate the target class of trojanned models. We evaluate the performance of the proposed framework on MNIST, NIST-Round0 and NIST-round1 dataset, and demonstrate the validness of our algorithm when the input are clean samples. In Addition, the classification capability of this algorithm is invariant of trigger types, trigger size when the trojanned model was trained since it only cares about the energy cost to move prediction to classifier boundaries.

%Ajmal's edit:
%\MS{it will be good to name this approach: like Achilles or Patroculus etc.} \RGnote{One possible name could be Cassandra, because in the Greek myth, Cassandra is the only one able to detect the soldiers inside the Trojan Horse} 
\vspace{-3mm}
Deep neural networks are being widely deployed for many critical tasks due to their high classification accuracy. In many cases, pre-trained models are sourced from vendors who may have disrupted the training pipeline to insert Trojan behaviors into the models. These malicious behaviors can be triggered at the adversary's will and hence, cause a serious threat to the widespread deployment of deep models. We propose a method to verify if a pre-trained model is Trojaned or benign. Our method captures fingerprints of neural networks in the form of adversarial perturbations learned from the network gradients. Inserting backdoors into a network alters its decision boundaries which are effectively encoded in their adversarial perturbations. We train a two stream network for Trojan detection from its global ($L_\infty$ and $L_2$ bounded) perturbations and the localized region of high energy within each perturbation. The former encodes decision boundaries of the network and latter encodes the unknown trigger shape. We also propose an anomaly detection method to identify the target class in a Trojaned network. Our methods are invariant to the trigger type, trigger size, training data and network architecture. We evaluate our methods on MNIST, NIST-Round0 and NIST-Round1 datasets, with up to 1,000 pre-trained models making this the largest study to date on Trojaned network detection, and achieve over 92\% detection accuracy to set the new state-of-the-art.
%it only cares about the energy cost to move prediction to classifier boundaries.
\end{abstract}

% \RGnote{according to neurIPS website, we need to add this: In order to provide a balanced perspective, authors are required to include a statement of the potential broader impact of their work, including its ethical aspects and future societal consequences. Authors should take care to discuss both positive and negative outcomes.}

\vspace{-5mm}
\section{Introduction}
\vspace{-3mm}
%Ajmal's edit 27/5/2020:
Deep neural networks (DNNs) are the main driving force behind the current success of Artificial Intelligence. However, training DNN models requires enormous amounts of data and computational resources. Hence, many users prefer to source and deploy pre-trained models in their, often security critical, applications such as drug discovery~\cite{chen2018rise,zhang2018seq3seq}, facial recognition~\cite{sun2015deepid3}, autonomous driving~\cite{geiger2012we}, surveillance~\cite{javed2002tracking}. It is well known that DNNs easily learn any bias that is present in the training data. Vendors of DNN models, with malicious intentions, can exploit this vulnerability of DNNs and intentionally inject Trojan behavior into the network during the training process. This is generally achieved by inserting a trigger into some of the  samples and then training the DNN to exhibit malicious behavior for data that contains the trigger and normal behavior for data without the trigger. With full control over the DNN training process, the adversary is able to choose any trigger shape. Triggers are chosen such that they do not appear suspicious to the human observer e.g. a yellow rectangular sticker on a stop sign can be used to trigger a DNN to classify it as a speed limit sign.  Since only the adversaries have knowledge of the trigger, they can initiate malicious behaviour at will and with no knowledge of the trigger, users of pre-trained models may not even suspect the presence of backdoors. This causes a serious threat to the widespread deployment of pre-trained models. Note that attacking Trojaned DNNs is much easier and different than adversarial attacks on clean DNNs, since the former has access to the DNN training process itself, while the latter only exploits intrinsic vulnerabilities of neural networks~\cite{akhtar2018threat}.

Given that noise-based adversarial attacks are inherent to CNN models, it is no surprise that trigger-based Trojan attacks also exist~\cite{yuan2019adversarial,tramer2017ensemble}. Trojans are generally inserted into the deep model during training or transfer learning~\cite{liu2020survey,liu2017trojaning,evtimov2017robust,chen2017targeted}. A backdoor is typically inserted into a network~\cite{gu2017badnets} %by adding malicious contents (triggers) to the training samples to compromise the training process and 
to make the CNNs mis-classify some specific class or classes. Instead of training a model with a dataset poisoned with triggers, another possible way the adversary can Trojan a network is by modifying the weights of selected neurons so that the model responds maliciously to a specific trigger \cite{liu2017trojaning}.  

Current challenges for Trojan (backdoor) detection in practice are: 1) Lack of a deep learning-based model trained on a large-scale dataset for Trojan detection; 2) Unavailability of trigger information for a suspected Trojan infected model; usually only limited training data of clean samples is available;  3) Very limited information which can be obtained from the query model predictions since the test accuracy for Trojaned DNNs is normal for clean inputs, and 4) The target class in the infected model is unknown, and it is computationally expensive to search all possible targeted attacks when the output labels are in the hundreds.

%Ajmal: the below para is already mentioned in literature review
%Neural Cleanse \cite{wang2019neural} is the first to propose a backdoor detection method for DNN models based on outlier detection of reversed triggers. DeepInspect \cite{chen2019deepinspect} applies conditional GNN to estimate triggers, and apply a similar outlier detection technique for the estimated triggers for Trojan detection, without the need for input image samples. However, these Trojan detectors are not learning-based, are far from being optimal and have not reported results on a large dataset. 

To address these challenges, we propose the first deep learning based Trojan Detection Network (TDN). Our method has two stages, the first one is a two stream neural network that outputs the probability of a model containing a Trojan, and the second stage predicts the target class in a Trojaned model. Our contributions are summarized as follows. First, we propose a deep neural network for Trojan detection from only a few clean samples. To the best of our knowledge, we are the first to use a DNN classifier, trained on a large scale dataset of benign and Trojaned models, for Trojan detection. Second, we propose a method for target class prediction in a Trojaned model. We introduce a new variable ($\gamma$) that quantifies the difficulty of attacking a model. This variable is a critical indicator for the target class of a Trojan infected model.% and achieve high classification accuracy for target class prediction.

{\bf Theoretical Justification:} Inserting Trojan behaviour into a network essentially puts an additional constraint on the model optimization during the training process. The model must learn to exhibit normal behavior and achieve an expected high classification accuracy on clean training/validation samples but exhibit the chosen malicious behaviour on samples containing a trigger, a localized pattern. This  has two important consequences. Firstly, the decision boundaries of the model must adjust to allow such a behavior. Secondly, the model must become more responsive to local patterns (the trigger). Our hypothesis is that if we can encode these two aspects, we will be able to detect Trojaned models accurately. For the former, we use universal adversarial perturbations \cite{moosavi2017universal} which, being image agnostic, reasonably capture a fingerprint of the decision boundaries. For the latter, we look for a localized region of high energy in the adversarial perturbation. Thirdly, we also hypothesize that Trojaned models are easier to fool with minimal universal perturbation energy compared to clean models. Our proposed method basically capitalizes on these three factors to detect Trojaned networks and the target class of such networks.

\vspace{-4mm}
\section{Related Work} %Ajmal: still needs polishing. I am skipping it
\vspace{-4mm}

%Ajmal: a sudden start with univ adv perturbations doesn't look good here. the paper is about Trojans.
Adversarial attacks on CNNs have focused on the phenomenon of noise-based adversarial examples~\cite{szegedy2013intriguing,akhtar2018defense}, which are visually almost indistinct from the original images, but can mislead DNN classifiers into making incorrect predictions. Even universal adversarial perturbations~\cite{moosavi2017universal} have been discovered that are image agnostic and when added to any image of any class, can cause the DNN to mis-classify them. By computing singular vectors of the Jacobian matrices of hidden layers, universal perturbations can be constructed with very few images~\cite{Khrulkov_2018_CVPR}. Adversarial attacks generally do not assume access to the training process of deep models. A comprehensive survey of such method is reported in \cite{akhtar2018threat}. In this paper, we focus on defending against Trojan attacks where the attacker disrupts the training pipeline of the DNN to insert a backdoor.

The risk of Trojan models arises when the training process of a DNN is outsourced or a pre-trained model from an untrusted source is deployed. This security risk was first investigated in Badnets\cite{gu2017badnets}. It was shown that backdoors in networks infected with Trojans can remain a threat even after transfer learning. Chen et al.~\cite{chen2017targeted} proposed a backdoor attack algorithm that uses poisoned data to contaminate the CNN model. Trojaning attack~\cite{liu2017trojaning} introduced a way to generate triggers and maximize the activation of some specific neurons to insert a backdoor. The embedded backdoors are stealthy and the unexpected malicious behavior is activated only by triggers, making them extremely challenging to detect with only clean data samples. 

Defense methods were first developed to detect adversarial images ~\cite{Akhtar_2018_CVPR,xie2019feature,yuan2019adversarial,Liao_2018_CVPR,grosse2017statistical,hendrycks2016early}. Metzen et al.~\cite{metzen2017detecting} detect  adversarial perturbations with a target classification network. Feinman et al.~\cite{feinman2017detecting} also use a binary classifier to detect adversarial perturbations. Magnet~\cite{meng2017magnet} trains a classifier on manifolds of normal examples to discriminate adversarial perturbations without any prior knowledge of the attack. Safetynet~\cite{lusafetynet} is designed to detect adversarial-noise based attacks and exploits the different adversarial perturbations produced to train a SVM classifier.

% [12] Y. Gao, C. Xu, D. Wang, S. Chen, D. Ranasinghe, S. Nepal, “STRIP: A Defence Against Trojan Attacks on Deep Neural Networks”, CoRR, arXiv:1902.06531, Feb 2019.

Methods for detecting and defending against Trojan attacks have also been proposed. Liu et al. ~\cite{liu2018fine} proposed a pruning and fine-tuning procedure to suppress backdoor attacks.  Chen et al. \cite{chen2018detecting} proposed Activation Clustering methodology for detecting and removing backdoors from DNNs. SentiNet\cite{chou2018sentinet} uses the behavior of adversarial misclassification of poisoned networks to detect an attack. However, all these methods fail in the realistic settings where access to poisoned data is not available.
Neural Cleanse~\cite{wang2019neural} was the first  method to detect Trojan infected models with clean samples by reverse engineering the trigger. They employ the Median Absolute Deviation (MAD) technique to  compute the anomaly in the $L_1$ norm of the reversed triggers to detect Trojaned models. However,  the trigger must be reverse engineered  for each class, which is not scalable in practice for DNNs with hundreds and thousands of classes.  DeepInspect~\cite{chen2019deepinspect} uses conditional GAN to reconstruct trigger patterns for Trojan detection.
NeuronInspect~\cite{huang2019neuroninspect} detects backdoor from the output features, such as sparsity, smoothness, and persistence of saliency maps obtained from back-propagation of the confidence scores. 
 Tabor~\cite{guo2019tabor} propose metrics to measure the quality of reversed triggers  and achieve improved performance than Neural Cleanse by introducing several regularization terms to refine the generated triggers. 
%  However, this comes at the cost of expensive computations.

The above methods~\cite{wang2019neural,chen2019deepinspect,huang2019neuroninspect} 
are sub-optimal because they are not learning-based and employ the MAD technique and manually tuned anomaly thresholds to detect the outliers of reverse engineered triggers. More importantly, none of these techniques report results on large scale data of benign/Trojaned models and none of them can predict the target class of a Trojaned model. To address these challenges, we propose {\it Cassandra}, a Trojan detection method that exploits universal adversarial perturbations~\cite{Khrulkov_2018_CVPR} generated from a very limited number of clean samples. Given their image-agnostic nature, we compute universal adversarial perturbations for a batch of clean samples, where the batches could be as few as 5. Note that this holds even if the number of classes is in thousands, unlike prior work such as Neural Cleanse, where one perturbation per class is necessary. Our method also provides the target class of a Trojan infected model.

% \section{Approach}
\vspace{-4mm}
\section{Detecting Trojan Infected Models}
\label{sec:blind}
\vspace{-3mm}

\begin{figure}
\centering
\includegraphics[width =\linewidth, trim=2cm 0cm 0cm 0cm, clip]{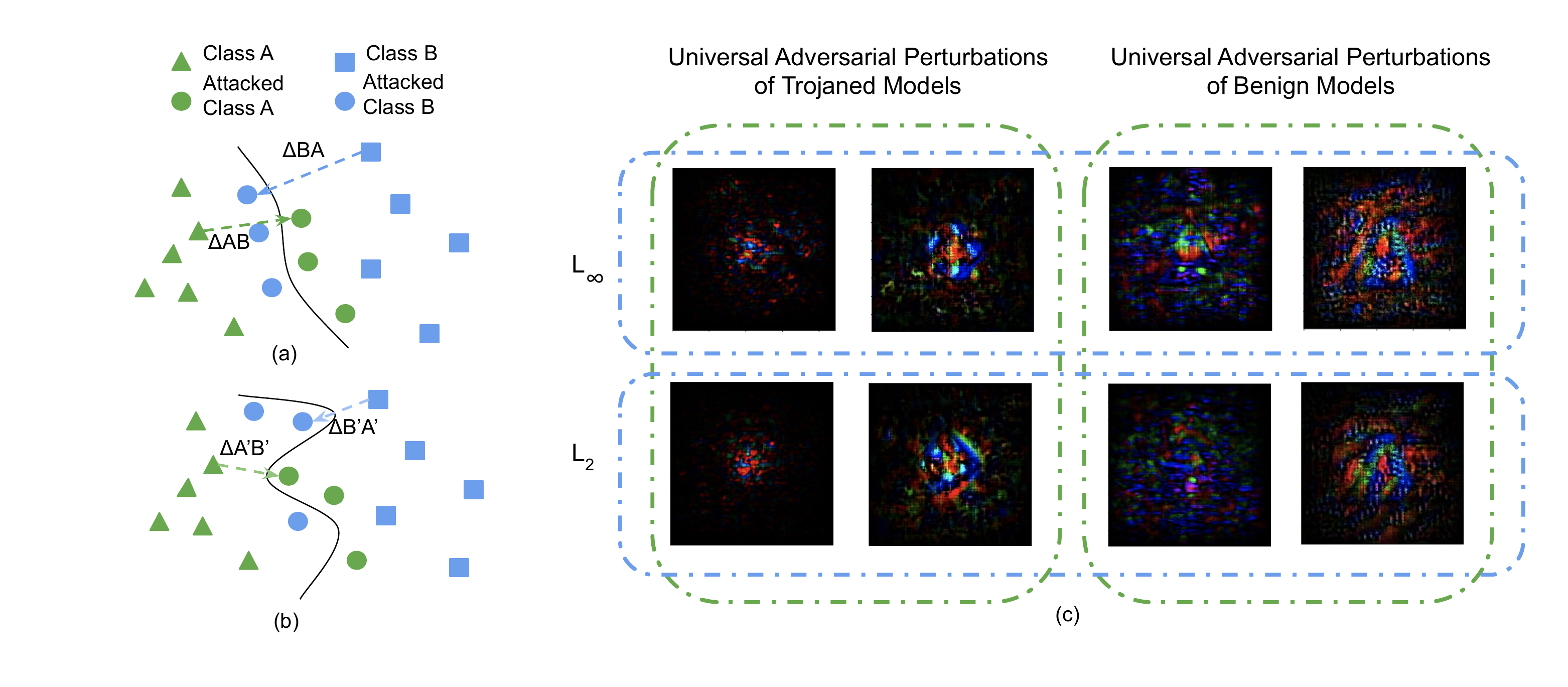}
\vspace{-10mm}
\caption{(a) and (b) illustrate how decision boundaries can change after inserting a Trojan in a network. The Trojaned model (b) has a complicated decision boundary after being compromised.
Introducing triggers in the training data changes the decision boundary of model (b) to accommodate the poisoned samples. This makes it easier to perturb the class label of a sample from class A to B (shown by arrows) since the distance across decision boundary is smaller compared to in (a). (c) Shows universal adversarial perturbations computed using $L_{\infty}$ (top row) and $L_{2}$ norm (bottom row) for benign and Trojaned models. Models from left to right: Trojaned Inception-v3, Trojaned DenseNet-121, benign ResNet50, and benign DenseNet121.}
\label{fig:db}
\vspace{-4mm}
\end{figure}

% \RGnote{Shouldn't there be more explanation about how decision boundaries are connected to adversarial perturbations ? My understanding is that given that Trojan attacks aim to modify the network such that it misclassifies examples for various classes as the target class, it would reorient decision boundaries to be such that there exist points that are close to the region of every class in the dataset (kind of like tripoints in international borders, where borders of multiple countries meet: https://en.wikipedia.org/wiki/Tripoint).}

% \RGnote{will work on improving the approach introduction based on content from later on in the section}
%We aim at retrieving the fingerprint to the backdoor inserted by the adversary. The neural network can learn a decision boundary that partitions the underlying vector space into multiple subsets/classes according to the learned Representative features belong to each class e.g. projection AOB in Figure ~\ref{fig:db},  The output label of a classifier on the decision boundary is ambiguous.

%Ajmal's Edit:

%\RGnote{Rohit's edit, Xiaoyu please review for correctness}
%Ajmal: fingerprint doesn't sound correct

During training, a neural network simultaneously learns feature representation and decision boundaries that partition the feature vector space into the respective classes. %As illustrated in Figure ~\ref{fig:db}, more than one valid feature projections and a set of decision boundaries are possible for the same dataset.
When an adversary inserts a backdoor into a network, the decision boundaries are altered. Our hypothesis is that Trojan infected networks exhibit decision boundaries that are different from typical, benign classification networks. Our approach exploits this fact by retrieving the fingerprints of the decision boundaries of a network, and subsequently trains a classifier on these fingerprints to classify a query network as benign or Trojan infected. We use adversarial perturbations to retrieve fingerprints of the decision boundaries of the query network. In contrast to image specific perturbations, universal perturbations~\cite{moosavi2017universal} are image agnostic,  such that the generated perturbations when added to any input image sends it across the decision boundary to change its label. The success of universal perturbations is measured by its fooling rate, the  proportion of images that are successfully mis-classified after the perturbation is added. Since universal adversarial perturbations capture the geometry of the decision boundaries~\cite{moosavi2017universal}, the perturbations for benign and Trojan models are expected to be significantly different in character. %Our Trojan detection network relies on classifying these perturbations. 

% through the back-propagation, that is concealed in the universal perturbations. 
 %Ajmal: not sure what you mean by AOB?
%The output label of a classifier on the decision boundary is ambiguous. %Ajmal: how is the previous sentence relevant?
% Our proposed approach aims to retrieve fingerprints of Trojans intentionally inserted into a neural network by an adversary during training, using poisoned data augmented with trigger patterns, 

\vspace{-2mm}
\subsection{Fingerprinting Decision Boundaries with Adversarial Perturbations}
\vspace{-2mm}

% It has been demonstrated in BadNets \cite{gu2017badnets} that adversarial attacks essentially exploit geometric correlations between the model decision boundaries and adversarial manipulations. In this paper, we exploit a similar approach to discover  fingerprints of a query network in terms  of patterns that encode model-specific decision boundaries.

%Thus perturbation energy and fooling rate 
%are prominent indicators for distinguishing infected models from clean ones. 

We formulate Trojan detection as a classification problem. For a query neural network model, $f$,  we define a  Trojan detection classifier $F$ as

\vspace{-3mm}
\begin{equation}\label{eq:tc}
F (\triangle x(f), E(\triangle x(f)), \eta(f)) = \left\{ \begin {array}{ll}
0 & \mbox{for benign model}\\
1 & \mbox {for Trojan infected model.}
\end {array}
\right.
\end {equation}
\vspace{-2mm}

%denotes a distribution of data samples
%in ${\Bbb R}^{d}$ and 

% And let $\mathbb{P}$ denotes a probability distribution of output prediction $y$ of classifier,  with mean  $\mu \in {\Bbb R}^{d}$, 
% such that for some fooling rate $\eta$  and  $||\triangle x||_p \le \epsilon $

 Here $f$ outputs a prediction $y$  for each input image $x$, drawn from the distribution $\mu$ of images in $\mathbb{R}^{d}$, $\eta(f)$ is fooling rate and $E(\triangle x(f))$ is the perturbation energy. Similarly, $\hat{f}$ and $\hat{y}$ denote the Trojaned classifier and corresponding prediction for $x$. For a desired threshold $\delta$, we  obtain universal perturbations $\triangle x$ and $\triangle \hat{x}$ for classifiers $f$ and $\hat{f}$, respectively, such that the following holds:
\begin{equation}\label{eq:cf1}
\underset{x \sim \mu}{\mathbb{P}}\{f(x +  \triangle x) \neq y \} \ge 1- \delta \ \quad \text{and} \quad
\underset{x \sim \mu}{\mathbb{P}}\{\hat{f}(x +  \triangle \hat{x}) \neq \hat{y}\} \ge 1-\delta \ ,%{\rm for~some~fooling~rate~} \eta
\end{equation}

Note that the observed fooling rate $\eta$ can go much higher than $1-\delta$ during the generation of universal adversarial perturbations. We define the perturbation energy $E$ as:
\begin{equation}\label{eq:ef}
E = \norm{\triangle x}_1 = \norm{h(x)}_1,
\end{equation}  
where $h$ is parameterized by the process that generates the perturbations. Let $E_{\triangle BA}$ denote the perturbation energy cost to transform all data samples from class B to class A across the decision boundary for a benign model, and vice versa for $E_{\triangle AB}$. Similarly, $E_{\triangle A'B'}$ and $E_{\triangle B'A'}$ denote the same for a Trojan infected model. In an infected model the decision boundary is changed such that some backdoors are created close to other classes. Due to these changes in the decision boundary, $E_{\triangle A'B'} < E_{\triangle AB}$ and $E_{\triangle B'A'} < E_{\triangle BA}$ for a given fooling rate (see Fig.~\ref{fig:db}a,b), where $\triangle AB$ is proportional to $E_{\triangle AB}$ and so on.

% where $\triangle AB$,$\triangle BA$,$\triangle A'B'$, and $\triangle B'A'$ are respectively proportional to $E_{\triangle AB}$,$E_{\triangle BA}$, $E_{\triangle A'B'}$  and $E_{\triangle AB}$. 

% The overall energy of the perturbation can be averaged over the image samples for classes A ($N_{A}$) and B ($N_{B}$).

% \MS{Fix equation}
% \vspace{-4mm}
% \begin{equation}
% \label{eq:eab}
% E_{\triangle AB} = \underset{\forall x_i \in A}\sum{E_{\triangle AB}(\triangle x_i)} \quad \text{and} \quad  E_{\triangle A'B'} = \underset{\forall x_i \in A}\sum{E_{\triangle A'B'}(\triangle x_i)}
% \end{equation}
% \vspace{-4mm}

% \begin{equation}
% \label{eq:eba}
% E_{\triangle A'B'} = \underset{0<i<N_{A}, 0<j<N_{B}}\sum{E_{\triangle B'_{j}A'_{i}}}
% \end{equation}

We define the notion of attack difficulty for both universal perturbations and targeted attacks as 
\begin{equation}\label{eq:defineAD}
\gamma = E / S,
\end{equation}
where $S$ is the fooling rate ($\eta$) for universal perturbations and attack success rate for targeted attacks. Universal adversarial perturbations of clean and Trojan infected models are distinguishable above a given fooling rate $\eta$, both visually and in terms of energy, as shown in Figure \ref{fig:db}c.
% its reciprocal $1/\gamma$ can be called attack efficiency.

% \begin{figure}
% \centering
% \includegraphics[width = 0.8\columnwidth]{figs/pert.png}
% \caption{Universal Perturbations for clean and trojaned models
% }
% \label{fig:pert}
% \end{figure}

%Our recent work %Ajmal: cannot say 'our' in blind reveiw

% \RGnote{Shouldn't some of this information (universal perturbation, perturbation energy?) be added to related work as well ? }
% \XZnote{Yes, but here we defined perturbation energy and also attach difficulty or its reciprocal : attack efficiency} \\

%F (\triangle x(f), E(f), \eta(f)) = \bigg \{  0 {\rm ~for~clean~model},  1 {\rm ~for~trojaned~model} \bigg \}

%\end{equation}

\vspace{-4mm}
%\section{System Architecture}
\section{Trojan Detector}
\vspace{-3mm}

\begin{figure}[t]
\centering
\includegraphics[width=\linewidth]{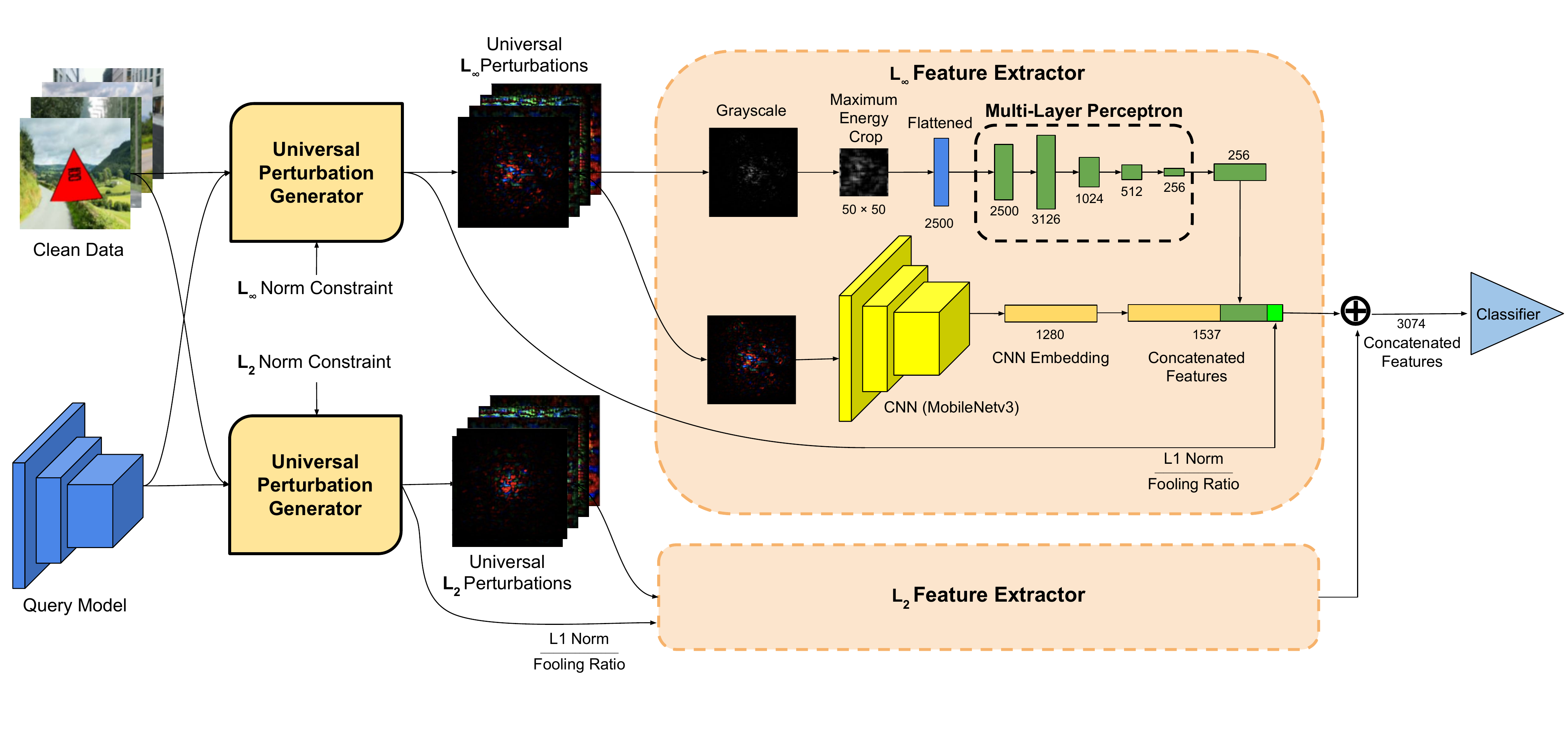}
\vspace{-12mm}
\caption{\textbf{Trojan Detection:} Features extracted from universal adversarial perturbations are used to fingerprint the query model. A two stream architecture is used with $L_2$ and $L_\infty$ norm bounded perturbations. The two feature extractor modules have identical architectures, but don't share weights. Three sets of features characterizing the query model are extracted: features to represent the maximum energy window in the perturbation extracted by a 5 layer MLP, CNN embedding from the perturbation image, using a MobileNetv3 network and the attack difficulty ($\gamma$) of the generated perturbation. Features from the two streams are concatenated and passed to the fully connected classification layer.}
\label{fig:system}
\vspace{-4mm}
\end{figure}

Figure~\ref{fig:system} shows the schematic overview of our proposed Trojan detector, referred to as \textit{Cassandra}. The query network, along with the clean labelled training data, are used to generate two types of universal perturbations i.e. those bounded by $L_\infty$ and $L_2$ norms. %Ajmal: what thresholds are used for two perturbations?
 {\it Note that we do not assume the presence of triggered images in the training data,  since triggers are unknown in a realistic scenario. }
The $L_\infty$ universal perturbations are fed to one stream of the network together with their corresponding attack difficulty, $\gamma_\infty$. Similarly, the $L_2$ norm bounded universal perturbations and their attack difficulty, $\gamma_2$, are fed to the second stream of the trojan detection network. The feature extractor in Fig.~\ref{fig:system} extracts distinguish features from the bounded perturbations of each stream. The two feature extractors have identical architectures, but have different sets of weights. Outputs from the two streams are concatenated and used to train the Trojan classifier with binary cross-entropy loss.

 %Ajmal: need to define 'windowed' and fooling rate at 'what'

% $\gamma$ = E / ($S$+$\epsilon$), where $\epsilon$ is a very small coefficient.

% The backbone stream of our TDN is shown in Figure~\ref{fig:bs} which  mainly contains a perturbation generator, feature extractor, feature concatenator\XZnote{where is it in the figure?} and a classifier. The other stream has a similar structure and feeds into the same feature concatenator.

\textbf{Perturbation Generator:} Since the target class of the (potentially Trojan infected) query network is unknown, the  universal adversarial perturbations (Eq.~\ref{eq:cf1}) \cite{moosavi2017universal} are computed  that cause mis-classification of any input image. The DeepFool \cite{moosavi2016deepfool} kernel is used  for perturbation generation. A batch of training images are passed to the query network, the direction of the nearest decision boundary is computed which is back-propagated to compute a small $L_\infty$ or $L_2$ bounded perturbation for the input. By iteratively refining the perturbation over different mini-batches, a universal (image agnostic) adversarial perturbation is obtained. The generated perturbations are sent to their respective feature extractor stream for further processing. In our case, we stop the iterations when a certain threshold $1 - \delta$ is achieved by the universal perturbation or a maximum number of iterations are reached.

%It takes the predictions from the Query model and then back-propagate the predictions and calculate perturbations to minimize the cost, the final obtained universal perturbations stacked with the original image will make the model give wrong prediction once it was fed to the model again. 

% Assume $\mu \in {\Bbb R}^{d}$ denotes a distribution of data samples in ${\Bbb R}^{d}$, $f$ define a classification function\XZnote{duplicate to equation 1?}
% \begin{equation}\label{eq:}
% f(x +  \triangle x) \neq f(x)  \ ,  {\rm~for~most~} x \in  \ \mu
% \end{equation}  
% subject to following  two conditions: 
% 1)  $||\triangle x||_p \le \epsilon$ ($p=\infty$ or $p=2$), and 
% 2) $\mathbb{P}(f(x + \triangle x) \neq f(x))$ \ge $\eta $, 
% %Ajmal: I have fixed many notations above. At least see the equation after compiling to ensure it is what you wanted it to be. Also, wasn't \triangle x the perturbation in previous text? And now is it v or \triangle v? yes, you are right
% where   $\epsilon$ controls the magnitude of the perturbation vector  $v$, and $\eta $ quantifies the desired fooling rate for all images sampled from the distribution $\mu$

\textbf{Feature Extractor} contains two parallel modules. The first one (top right in Fig.~\ref{fig:system}) is a Multi Layer Perceptron (MLP), which crops a $50\times 50$ window from the perturbation image (after conversion to grayscale) and outputs a $256$ dimensional feature vector. The MLP layers have $2500$, $3126$, $2048$, $1024$,  $512$ and $256$ neurons, respectively. 
%The location of the window used an input to the MLP is decided based on the maximum energy. 
A sliding window is moved over the grayscale perturbation and the location which has the maximum $L_1$ norm is selected as input to the MLP. The second module (bottom right in Fig.~\ref{fig:system}) comprises of a MobilenetV3-Large CNN \cite{howard2019searching} pre-trained on ImageNet classification. The 1280-dimensional embedding output from the penultimate layer is used as a feature for the Trojan classifier. The network has a total of 5.5M trainable parameters.

%The $224 \times 224 \times 3$ input image is initially fed to Conv $3 \times 3$, feature maps = 16; followed by 15 bottleneck blocks, where each block consists of 3 convolution layers and one SqueezeExcite (2 convolution layers) block, resulting in 160D feature maps. The feature maps of the last bottleneck are processed by Conv $1 \times 1$, to generate a $7 \times 7 \times 960$ volume, which  are then fed to a pooling layer with filter size $7 \times 7$, the output volume size become $1 \times 1 \times 960$, and then are convolved with 1 Conv $1 \times 1$ filter, the final output size is 1280. Both ReLU and h-swish nonlinearity are used in the network.

\textbf{Trojan Classifier:} The output of each feature extractor module is a concatenation of the MLP features (256-D), CNN features (1280-D) and the attack difficulty (1-D). This totals to a 1537-D vector. The outputs of the two-streams ($L_\infty$ and $L_2$ perturbations) are concatenated to form a 3,074-D vector that is fed to the Trojan classifier, which is a simple fully connected layer. The probability of  the query model being  Trojan infected is obtained by applying the sigmoid activation to the output. To fully capture properties of the complex decision boundaries, we divide the training data into 10 batches and obtain 10 probabilities for each query model. The final score is computed as the mean value of these 10 probabilities. 
%  Our Trojan classifier has two ouputs, one is 0 or 1 for classification task, which is the argmax of the final softmax scores, the other one is a real between $0$ and $1$ which indicates Trojan Probability of the query network being Trajaned or not.

\vspace{-4mm}
\section{Target Class Prediction}
\vspace{-4mm}

% \RGnote{(To Self) There's redundancy between algorithm table and this section. Need to revise again.}

%Ajmal: refering to NeuralCleanse here is not relevant as it is for a different purpose and using a different method
%NeuralCleanse ~\cite{wang2019neural} performs outlier detection using the $L_1$ norm of reversed triggers (perturbations) for each class to detect Trojans which is computationally expensive. %We use a similar approach only for the target class prediction once we have already detected a Trojan infected model. Moreover, instead of the $L_1$ norm of the reversed trigger, we propose a new metric i.e. attack difficulty ($\gamma$) for outlier detection.  %propose a new variable to perform outlier detection for target class, called attack difficulty 

%Ajmal: my edit
We propose targeted attack difficulty, as a metric for outlier class prediction in a Torjan infected model. An outlier class is the one which is easy to launch a targeted attack against, compared to the other classes, and hence most likely to be the target class of the Trojan infected model.

\begin{figure}[t]
\centering     %%% not \center
\subfigure{\includegraphics[width=.45\textwidth]{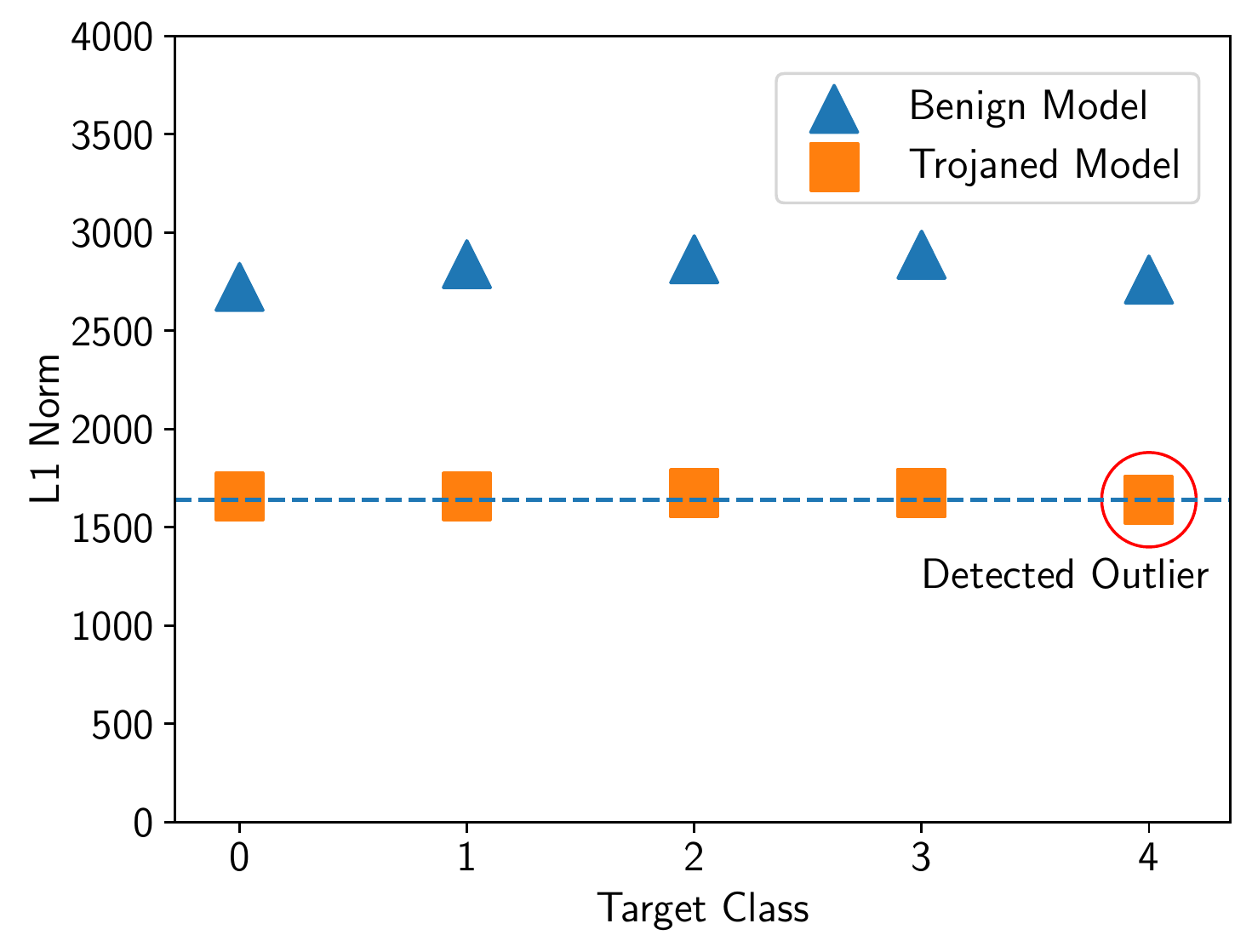}}
\subfigure{\includegraphics[width=.45\textwidth]{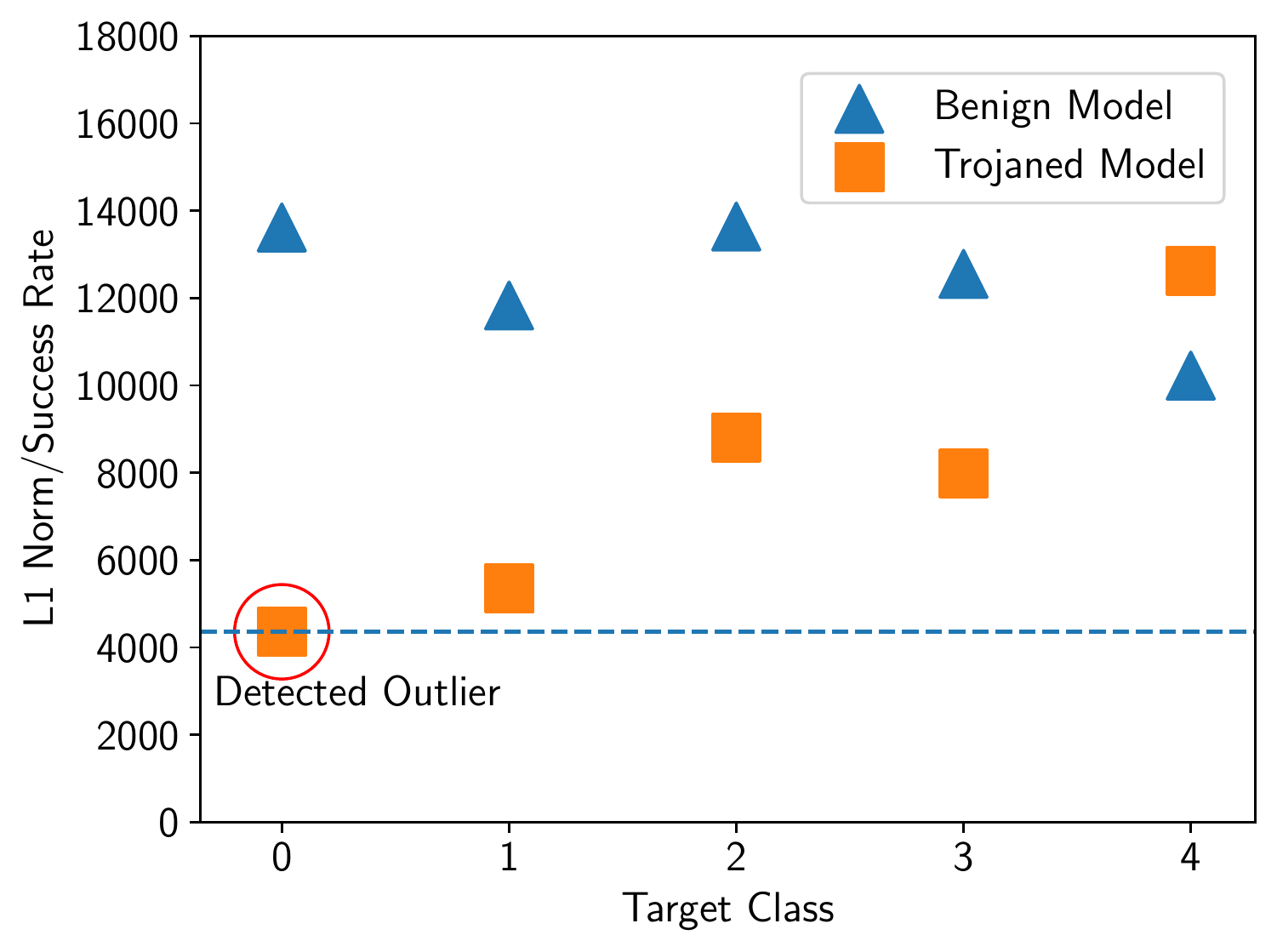}}
\vspace{-3mm}
  \caption{\textbf{Target class prediction:} The $L_1$ norm (left) and attack difficulty ($L_1$ Norm/Success Rate) (right) of FGSM adversarial perturbations computed per label for benign and Trojan infected networks. The true target class is 0, detected outlier in each case is encircled.}
\vspace{-3mm}
\label{fig:plots}
\end{figure}

Attack difficulty is defined as $\gamma = E/S$,  where $E$ is the perturbation energy (Eq.~\ref{eq:ef}) and $S$ is the attack success rate for the targeted attack i.e. the proportion of images whose predictions change to target labels. Attack difficulty (or its reciprocal attack efficiency) measures the perturbation energy normalized by the success rate of the attack. Given a query model, we use the Fast Gradient Sign Method (FGSM)~\cite{goodfellow2014explaining}, given its fast execution time, to compute adversarial perturbations for each class. For example, NIST-Round0 data contains five class labels 
%If all of these classes are mis-classified as class 2 for a Trojan model, class 2 is the target class of the Trojan attack and  0, 1, 2, 3 and 4 are called the triggered class. 
and the Trojan models classify triggered images of any class to class 0. In this case, class 0 is the target class and the attack is called an "any-to-one" targeted attack.
Fig.~\ref{fig:plots} shows that the proposed attack difficulty is able to correctly detect the target class of the Trojan attack as outlier, but $L_1$ norm used for Trojaned model detection in Neural Cleanse~\cite{wang2019neural} fails. We finalize our target class prediction with a two stage method. The first stage is our Trojan detection network which outputs the probability of the model being infected with a Trojan, and the second stage is outlier detection based on Median Absolute Deviation (MAD)~\cite{hampel1974influence,wang2019neural} for predicting the target class. The anomaly index for outlier detection is defined as the absolute deviation of the data points from their median and then  normalized by the median, to measure the dispersion of the data distribution. For Trojaned models, the second stage selects the label with anomaly index value above a threshold as the predicted target class.

% The models used are DenseNet models from the NIST Round-0 Dataset
% \RGnote{Change predicted class to "outlier"}

\vspace{-4mm}
\section{Experiments}
\vspace{-4mm}

%Ajmal: no need to have this subsection title
%\subsection{Implementation Details} 
%Except otherwise stated, %Ajmal: but you haven't stated anywhere
For all experiments, we perform 5-fold cross validation and report average results.  In Eqn.~\ref{eq:cf1}, $\delta$ is set to 0.2 to quantify the desired fooling rate.  We use the Adam optimizer with a learning rate of 0.001.The constant estimator for the MAD outlier detector is 1.4826, so that any data sample with anomaly index larger than 2 has $>$ 95\% probability of being an outlier. We employ anomaly index threshold of 2, such that the class labels with anomaly index larger than 2 are considered the target class. For training we use a server with 6 Nvidia RTX 2080 Ti GPUs. Perturbation generation and training for NIST-Round1 data takes around 12 hours. Inference for each model takes about 560s with Nvidia RTX 2080 Ti, and inference for 200 models finishes within 24 hours. 

% In this section, we first detail the experimental setup such as dataset description and trojan model generation etc. Then we report the performance of the our trojan detector for MNIST, NIST-Round0 and NIST-Round1 tasks as state-of-the-art methods. Further more, we report the results of our target class prediction. Finally, we provide the ablation study for the sensitivity of the hyper-parameters of our trojan detector, the threshhold parameter e.g., the number of iteration and magintude of perturbation $\epsilon$ for universal perturbation generation.

\vspace{-3mm}
\subsection{Datasets}
\vspace{-2mm}

We evaluate our proposed approach on a dataset of trigger infected models for classifying images from MNIST and the public NIST-Round0 and NIST-Round1 datasets. We will refer to the dataset of trigger infected MNIST classification models as Triggered MNIST dataset throughout. Code to generate the triggered MNIST dataset was used from the TrojAI GitHub repo\footnote{\url{https://github.com/trojai}}. NIST datasets were obtained from the TrojAI challenge website\footnote{\url{https://pages.NIST.gov/trojai/docs/data.html\#download-links}}.
% \begin{itemize}
% \item 

\begin{figure}[h!]
\centering
\includegraphics[width=0.9\columnwidth]{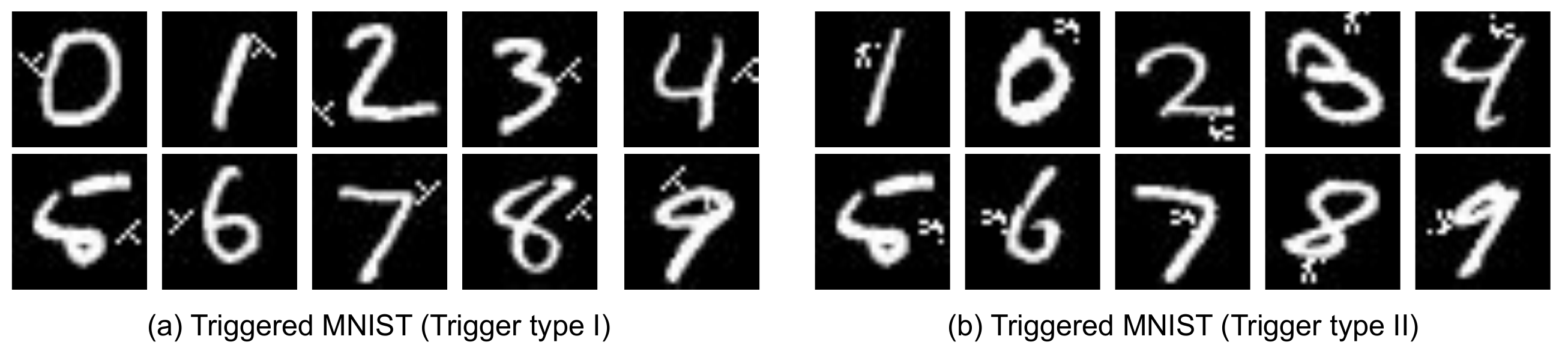}
\vspace{-3mm}
\caption{Triggered MNIST dataset samples containing Type I triggers (a) and Type II triggers (b).}
\label{fig:trigger}
\vspace{-3mm}
\end{figure}

\textbf{Triggered MNIST Dataset:}
Two types of triggers, Type I and II (see Fig.~\ref{fig:trigger}) are inserted into each image of clean MNIST dataset to generate Triggered data. A total of 900 models of 3 architectures (ModdedBadNet, BadNet and ModdedLeNet5) are generated. Out of these, 300 were benign models, 300 were trained for any-to-any attack, and 300 were trained for any-to-one targeted attack. Details of the models and their performance on clean and triggered data are given in the supplementary material.

\begin{figure}[h!]
\centering
\includegraphics[width=0.9\linewidth]{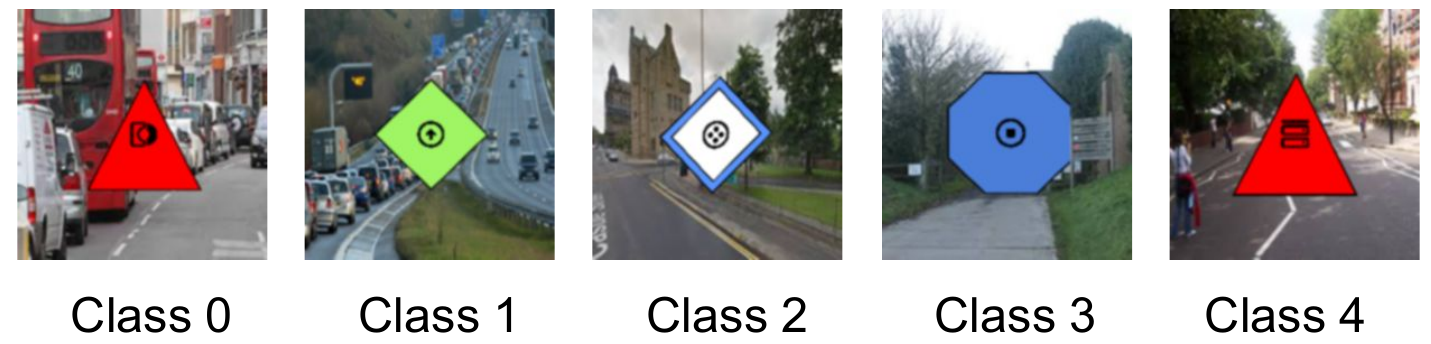}
\vspace{-3mm}
\caption{Clean image samples of 5 classes from the NIST datasets.}
\label{fig:NISTdata}
\vspace{-3mm}
\end{figure}

\textbf{NIST Datasets:}
The NIST datasets consist of traffic sign classification models (half benign and half Trojaned) with 3 possible architectures (Inception-v3, DenseNet-121, and ResNet50). The models were trained on synthetically created images of artificial traffic signs superimposed on road background scenes. The Trojan infected models are poisoned with an unknown embedded trigger. \textbf{NIST-Round0} and \textbf{NIST-Round1} datasets are both from the same distribution, the main difference is that Round0 consists of 200 models, while Round1 dataset has 1,000 models. Details of the  models in the NIST datasets including their accuracy and attack success rates are provided in the supplementary material. Clean data samples used to train the NIST models can be seen in Figure~\ref{fig:NISTdata}.

% \end{itemize}
% \begin{itemize}%Ajmal: why so many indents when you are short of space
% \item \textbf{MNIST and Triggered MNIST Dataset}
%\begin{itemize}
    % The MNIST dataset has 10 classes with  70,000 clean images (without triggers). 

% The success rate and classification accuracy for both clean and trojaned models are shown in Table~\ref{tab:NIST0_model}

% \item \textbf{NIST-Round1}

% \end{itemize}

% \begin{table}[h]
%         \centering
%         \caption{Attack success rate and classification accuracy for four types of trojaned models on NIST-Round1 datasets.Success rate is the proportion of images that change to target labels for trojaned model.}
%         \begin{tabular}{c|c|c|c|c|c}
%         \hline
%         %\abovespace\belowspace
%         Model Type &  \multicolumn{4}{c}{Trojan Model} & benign model \\
%         \hline

%          & \multicolumn{2}{c}{Attack Success Rate} & \multicolumn{2}{c}{Classification Accuracy} & Classification Accuracy \\
%         \hline
%           & &  & &  &  \\
%         \hline\hline
%         Densenet121  & &   &  &  & \\
%         inceptionv3 &99.54 &  & 99.63 & & \\
%         resnet50 & & &  &  & \\
%         % Resnet18 & 98.5 &98.8  & 99.1 &98.9 & 99.2 \\
%         \hline
%     \end{tabular}
%         \label{tab:NIST0_model}%
% \end{table}

\vspace{-3mm}
\subsection{Results}
\vspace{-2mm}

% Our proposed algorithm is invariant to any trigger specific information such as  trigger type, color or size.  It also does not require tirggered images. This is a realistic setting, since the knowledge of the trigger and sample triggered images are unlikely to be available. The proposed TDN can be trained with Trojanned models that are generated with any type of triggers.
In Table~\ref{table:model-table7}, we report the classification accuracy for a variety of training and test set model configurations for the Triggered MNIST dataset. The classification accuracy is consistently  high for both Type I (93.3\%) and Type II (91.7\%) triggers. %, training and test sets only contain infected models of trigger type I(or trigger type II),  . 
Even when training and test models are infected by different types of triggers, %type I(or II) trigger and test models that are embedded with trigger type II(or I), or vice versa, 
the algorithm still has a high classification accuracy of 91.7\% (or 90\%) which shows that our method is independent of trigger types. We achieve 94.4\% performance for the configuration where both  trigger types are present, in equal proportions, in the training and test sets (Table \ref{table:model-table7} last row). %\MS{not really clear}

\begin{table}[h!]
	\centering
	\caption{\textbf{Trojan Detection results} for the Triggered MNIST dataset of models. The proposed method achieves good results even when trained on a trigger, which is  different  from the one seen at test time.}
	\footnotesize
%	\begin{adjustbox}{width=0.9\linewidth,height=20 \lineheight, center}
	
		\begin{tabular}{|c|c|c|c|c|c|}
 		\hline
%    Input Samples & Training Set models & num of Training set models & Testing Set models & num of test set models & Classification Accuracy. \\
\# \textbf{Train Models}  & \textbf{Train Trigger} & \# \textbf{Test Models}  & \textbf{Test Trigger} & \textbf{Accuracy} \\
\hline \\[-1em] \hline
 240  & I     & 60  & I     & $93.3\pm1.6$ \\
        \hline
 240  & II    & 60  & II    & $91.7\pm1.7$ \\
        \hline
  240  & I     & 60  & II    & $90.0\pm1.2$ \\
        \hline
  240  & II    & 60  & I     & $91.7\pm1.1$ \\
        \hline
  480  & I, II & 120 & I, II & $94.4\pm1.2$ \\
  \hline
%   any-to-one & 480  & I, II & 120 & I, II & $91.7\pm0.8$ \\
        % \hline
%   Any to one  & 480 & I,II,III & alpha 120 & I,II,III &\% \\
 		\hline
		\end{tabular}
%	\end{adjustbox}
	\label{table:model-table7}
\end{table}

\begin{table}[b!]
\caption{\textbf{Trojan detection accuracy on Triggered MNIST, NIST-Round0 \& Round1 datasets}.}
\begin{center}
\label{tab:comaprison_f}
\begin{tabular}{|l|c|c|c|}
\hline
 & \multicolumn{3}{c|}{\textbf{Dataset}} \\
\hline
\textbf{Method} & \textbf{Triggered MNIST} & \textbf{NIST-Round0}  &\textbf{NIST-Round1} \\ \hline
\\[-1em] \hline
Neural Cleanse~\cite{wang2019neural}& $76.6\pm1.2$  &$67.5\pm1.6$  & $74.0\pm1.4$  \\ \hline
\textbf{Cassandra (Ours)} & $94.4\pm1.2$ & $92.5\pm1.1$ & $92.0\pm1.3$ \\
\hline

\end{tabular}
\end{center}
\end{table}

NIST datasets are more challenging compared to Triggered MNIST, not only in terms of trigger types, color and size of the data used to train the infected models, but also due to the fact that the NIST models are much deeper. Our method obtains high classification accuracy of 92.5\% for NIST-Round0 and 92.0\% for NIST-Round1 datasets.  %Both NIST-Round0 and NIST-Round1 results show 
Table \ref{tab:comaprison_f} shows results of our method on the Triggered MNIST, NIST Round0 and NIST Round1 datasets and compares them to Neural Cleanse \cite{wang2019neural}. Our proposed Trojan Detection Network outperforms Neural Cleanse on all three datasets with large margins of $17.8\%$, $25\%$ and $18$\% respectively. This can be attributed to two reasons. Firstly, it is difficult for Neural Cleanse to find an optimal anomaly index threshold. Secondly, reverse engineering the trigger does not perform well when the triggers are complex.
% used in the NIST dataset. %Ajmal: we don't know how the NIST triggers look like + neural cleanse performs as good on NIST Round1 as it peforms on MNIST
\begin{table}[h!]
\caption{\textbf{Target Class Prediction Accuracy:} Availability of $P(Trojan)$ significantly increases the target class prediction accuracy. All models are attacked any-to-one i.e. one target class per model.}
\begin{center}
\label{tab:tar_pred}
\begin{tabular}{|l|c|c|c|c|}
\hline

 & \textbf{Triggered MNIST} &  \textbf{NIST-Round0} &
 \textbf{NIST-Round1}\\ \hline
%  &\textbf{Trojan Probability} &  &   & \\ \hline
\\[-1em] \hline
 without $P(Trojan)$ & $76.1\pm1.5$   & $72.5\pm2.5$ & $70.0\pm1.0$ \\  \hline
 with predicted $P(Trojan)$ & $90.0\pm1.0$  & $94.7\pm1.7$ & $88.1\pm0.7$ \\\hline
 with ground truth $P(Trojan)$ & $95.0\pm1.3$ & $98.8\pm1.4$ & $91.7\pm1.5$ \\ \hline

\end{tabular}
\end{center}
\end{table}

Table \ref{tab:tar_pred} shows our target class prediction results. The proposed two stage prediction algorithm based on the attack difficulty and predicted $P(Trojan)$ improves the classification accuracy significantly over the baseline (without $P(Trojan)$) from 76.1\%, 72.5\% and 70.0\% to 90.0\%, 94.7\% and 88.1\% on Triggered MNIST, NIST-Round0 and NIST-Round1 datasets respectively. Using Ground truth P(Trojan) further improves classification accuracy which demonstrates attack difficulty is a critical indicator of target class.

% We also test the transferability of our TDN between different types of query models. We observe that when the TDN was trained by one type of model architecture, it still provides acceptable classification accuracy~\ref{table:model-table7}. The classification accuracy drops because for different model architectures some spatial features keeps but the dynamic properties captured by MLP module is not consistent.\XZnote{leave it to discussion?}

% We compare the results of one stage outlier detection and two stage outlier detection for target class prediction. The one stage outlier prediction use MAD technique to predict trojaned models based on attack difficulty $\gamma$, if the prediction is trojaned model, the corresponding class label with the highest anomaly index is determined as the target class. Outlier detection based on MAD can detect multiple target class, since we only have one target class for our target class prediction datasets, we  consider the highest anomaly index.

% Target class predication results in Table~\ref{tab:tar_pred} shows the two stage strategy, with predicted Trojan Probability from the first Trojan detection stage, has 94.7\% classification accuracy for target class prediction, which is (22.2\%) higher compared to one stage prediction that is $72.5\%$. Similar results are observed for MNIST and NIST-Round1 data. If we know it's a trojaned model, the target class detection accuracy achieve 100\% for NIST-Round0, $95.0$ for NIST-Round1 and 91.7\% for MNIST data.

%\vspace{-3mm}
\subsection{Ablation Study}
\vspace{-2mm}

{\bf Trojan Detector Network Modules:}
In Table~\ref{tab:tro_det}, we explore different network architectures and the functionality of individual modules of our method. Using only universal perturbations computed from the complete training data of NIST-Round0, we achieved 77.5\% classification accuracy. After dividing the training data into 10 batches (these are different from the training mini-batches), we generate 10 perturbations for each model. With these 10 perturbations, the accuracy improves to 85\%. Adding the attack difficulty further improves the classification accuracy in all cases. Finally, with multi-batch and two stream architecture we achieve 92.5\% classification accuracy.

\begin{table}[h!]
\caption{\textbf{Effects of using multiple perturbations and attack difficulty.} Trojan detection accuracy on the NIST-Round0 validation data improves significantly after using multiple perturbations (n=10) calculated from different batches of training data. Using $L_\infty$ and $L_2$ perturbations in a two stream architecture combined with attack difficulty ($\gamma$) further improves the accuracy.}
\begin{center}
\label{tab:tro_det}
\begin{tabular}{|l|c|c|}
\hline
 & \multicolumn{2}{c|}{\textbf{Classification Accuracy}} \\ 
\hline
\textbf{Input for classifier} & \textbf{without $\gamma$}  & \textbf{with $\gamma$} \\
 \hline
 \hline 
% SVM & perturbation & 75.0  \\ \hline
$L_\infty$ perturbation from all training data & $77.5\pm 2.1$  & $82.5\pm 1.4$ \\ \hline
multi-batch $L_\infty$ perturbations                             & $85.0\pm1.8$ & $90.0\pm1.8$  \\ \hline
 multi-batch + two stream ($L_\infty$ \& $L_2$) perturbations  & $85.0\pm1.4$ & $92.5\pm1.1$  \\ \hline

\end{tabular}
\end{center}
\end{table}

\begin{table}[h!]
\caption{\textbf{Effect of perturbation generator hyper-parameters on Trojan detection accuracy}
}
\begin{center}
\label{tab:comap_hyper}
\begin{tabular}{|c|c|c|c|c|c|}
\hline
 & \multicolumn{5}{c|}{\rule{0pt}{1.2em}\textbf{ $L_{\infty}$ Perturbation Magnitude ($\xi_\infty/255$)}} \\ \hline
 & \multicolumn{5}{c|}{\rule{0pt}{1.2em}$\xi_{2}/255$ = 10, \# $L_2$ iterations = 10} \\ \hline
\textbf{$\textbf{\#}$ of Iterations} &\textbf{0.1} & \textbf{0.2}  & \textbf{0.4} & \textbf{0.8}  & \textbf{1}  \\ \hline
\\[-1em] \hline
 5 & $87.5\pm2.7$ & $87.5\pm1.4$  & $90.0\pm1.6$ & $90.0\pm1.4$  & $90.0\pm2.1$ \\ \hline
10 & $87.5\pm1.7$ & $90.5\pm1.1$  & $92.5\pm1.4$ & $92.5\pm1.2$  & $92.5\pm1.1$ \\ \hline
15 & $90.0\pm2.1$ & $90.5\pm1.3$  & $90.0\pm1.3$ & $92.5\pm2.1$  & $92.5\pm1.2$ \\ \hline
\hline
& \multicolumn{5}{c|}{\rule{0pt}{1.2em}\textbf{ $L_{2}$ Perturbation Magnitude ($\xi_2/255$)}} \\ \hline
& \multicolumn{5}{c|}{\rule{0pt}{1.2em}$\xi_{\infty}/255$ = 1, \# $L_\infty$ iterations = 10} \\ \hline
\textbf{$\textbf{\#}$ of Iterations} & \textbf{5}  & \textbf{10} & \textbf{20} & \textbf{30} & \textbf{40}   \\ \hline
\\[-1em] \hline
  5 & $87.5\pm1.8$ & $90.0\pm1.4$  & $90.0\pm1.8$ & $90.0\pm1.2$ & $90.0\pm1.4$ \\ \hline
 10 & $90.0\pm1.4$ & $92.5\pm1.1$  & $90.0\pm1.4$ & $92.0\pm1.8$ & $92.5\pm1.1$ \\ \hline
 15 & $90.0\pm1.2$ & $92.5\pm1.1$  & $92.5\pm1.2$ & $90.0\pm1.3$ & $92.5\pm1.2$ \\ \hline
\end{tabular}
\end{center}
\end{table}

% \vspace{-3mm}
% \subsection{Ablation Study on Hyper-parameters}
% \vspace{-3mm}

{\bf  Universal Perturbation Generator Hyper-parameters:} %In practice, the hyper-parameters used for universal adversarial perturbations among different tasks may be sensitive to data. 
The choice of hyper-parameters may impact on the effectiveness of the generated universal adversarial perturbations for various tasks. However, our experiments show that the proposed method is robust to these parameters. We compare the mean classification accuracy when using different number of iterations and magnitudes for $L_{2}$ and $L_{\infty}$ bounded universal adversarial perturbations, and find that the Trojan detection accuracy varies only slightly as shown in Table~\ref{tab:comap_hyper}.

\vspace{-3mm}
\section{Conclusion}
\vspace{-3mm}

We proposed the first deep learning based method, that is trained on a large scale dataset of Trojaned and clean models, for detecting Trojan infected models. 
%a novel strategy to detect Trojans intentionally embedded to the deep neural network by adversary. 
We exploit the universal adversarial perturbations to retrieve the fingerprints of Trojans in the DNNs and train our proposed TDN based on the features of the perturbations and attack difficulty to discriminate benign and Trojaned models. We also proposed simple variable, coined attack difficulty $\gamma$, to measure the energy needed to achieve an average unit fooling rate. Based on the attack difficulty, we proposed a two stage target class prediction method that can predict the target class of a Trojaned model in addition to the Trojan probability. This provides further information on the type of malicious behaviour embedded in a Trojan infected model e.g. which identity is being impersonated in a Trojaned face recognition model.

%  \newpage

% \clearpage

% \section{Broader Impact}
% \vspace{-3mm}

% \textbf{Positive impact}: Trojan attacks pose a potentially serious security threat and can cause dramatic loss of property or even lives. An adversary can create a neural network that has state-of-the-art performance on typical validation samples, but behaves maliciously on specific attacker-chosen inputs. A critical security problem to be solved before the pervasive deployment of deep learning models is to make sure that the models are free of Trojans (backdoors). We expect our work to set the new baseline for examining the integrity of deep models before their deployment. This will have a positive social impact on the axes of safety, security and accountability. 

% \textbf{Negative impact}: While we do not anticipate any direct negative impact from our research, it is possible that adversaries may take into account our research and improve the sophistication of their Trojan attacks in the future.

\section*{Acknowledgments}
This research  was supported in part under ARC Discovery Grant DP190102443. Xiaoyu Zhang and Rohit Gupta were supported by University of Central Floria ORC fellowships. 

\bibliographystyle{unsrt}
\bibliography{egbib}

\begin{thebibliography}{10}

\bibitem{chen2018rise}
Hongming Chen, Ola Engkvist, Yinhai Wang, Marcus Olivecrona, and Thomas
  Blaschke.
\newblock The rise of deep learning in drug discovery.
\newblock {\em Drug discovery today}, 23(6):1241--1250, 2018.

\bibitem{zhang2018seq3seq}
Xiaoyu Zhang, Sheng Wang, Feiyun Zhu, Zheng Xu, Yuhong Wang, and Junzhou Huang.
\newblock Seq3seq fingerprint: towards end-to-end semi-supervised deep drug
  discovery.
\newblock In {\em Proceedings of the 2018 ACM International Conference on
  Bioinformatics, Computational Biology, and Health Informatics}, pages
  404--413, 2018.

\bibitem{sun2015deepid3}
Yi~Sun, Ding Liang, Xiaogang Wang, and Xiaoou Tang.
\newblock Deepid3: Face recognition with very deep neural networks.
\newblock {\em arXiv preprint arXiv:1502.00873}, 2015.

\bibitem{geiger2012we}
Andreas Geiger, Philip Lenz, and Raquel Urtasun.
\newblock Are we ready for autonomous driving? the kitti vision benchmark
  suite.
\newblock In {\em 2012 IEEE Conference on Computer Vision and Pattern
  Recognition}, pages 3354--3361. IEEE, 2012.

\bibitem{javed2002tracking}
Omar Javed and Mubarak Shah.
\newblock Tracking and object classification for automated surveillance.
\newblock In {\em European Conference on Computer Vision}, pages 343--357.
  Springer, 2002.

\bibitem{akhtar2018threat}
Naveed Akhtar and Ajmal Mian.
\newblock Threat of adversarial attacks on deep learning in computer vision: A
  survey.
\newblock {\em IEEE Access}, 6:14410--14430, 2018.

\bibitem{yuan2019adversarial}
Xiaoyong Yuan, Pan He, Qile Zhu, and Xiaolin Li.
\newblock Adversarial examples: Attacks and defenses for deep learning.
\newblock {\em IEEE transactions on neural networks and learning systems},
  30(9):2805--2824, 2019.

\bibitem{tramer2017ensemble}
Florian Tramèr, Alexey Kurakin, Nicolas Papernot, Ian Goodfellow, Dan Boneh,
  and Patrick McDaniel.
\newblock Ensemble adversarial training: Attacks and defenses.
\newblock In {\em International Conference on Learning Representations}, 2018.

\bibitem{liu2020survey}
Yuntao Liu, Ankit Mondal, Abhishek Chakraborty, Michael Zuzak, Nina Jacobsen,
  Daniel Xing, and Ankur Srivastava.
\newblock A survey on neural trojans.
\newblock In {\em 2020 IEEE International Symposium on Quality Electronics
  Design (ISQED)}, 2020.

\bibitem{liu2017trojaning}
Yingqi Liu, Shiqing Ma, Yousra Aafer, Wen-Chuan Lee, Juan Zhai, Weihang Wang,
  and Xiangyu Zhang.
\newblock Trojaning attack on neural networks.
\newblock In {\em 25nd Annual Network and Distributed System Security
  Symposium, {NDSS} 2018, San Diego, California, USA, February 18-221, 2018}.
  The Internet Society, 2018.

\bibitem{evtimov2017robust}
Kevin Eykholt, Ivan Evtimov, Earlence Fernandes, Bo~Li, Amir Rahmati, Chaowei
  Xiao, Atul Prakash, Tadayoshi Kohno, and Dawn Song.
\newblock Robust physical-world attacks on deep learning visual classification.
\newblock In {\em Proceedings of the IEEE Conference on Computer Vision and
  Pattern Recognition}, pages 1625--1634, 2018.

\bibitem{chen2017targeted}
Xinyun Chen, Chang Liu, Bo~Li, Kimberly Lu, and Dawn Song.
\newblock Targeted backdoor attacks on deep learning systems using data
  poisoning.
\newblock {\em arXiv preprint arXiv:1712.05526}, 2017.

\bibitem{gu2017badnets}
T.~{Gu}, K.~{Liu}, B.~{Dolan-Gavitt}, and S.~{Garg}.
\newblock Badnets: Evaluating backdooring attacks on deep neural networks.
\newblock {\em IEEE Access}, 7:47230--47244, 2019.

\bibitem{moosavi2017universal}
Seyed-Mohsen Moosavi-Dezfooli, Alhussein Fawzi, Omar Fawzi, and Pascal
  Frossard.
\newblock Universal adversarial perturbations.
\newblock In {\em Proceedings of the IEEE conference on computer vision and
  pattern recognition}, pages 1765--1773, 2017.

\bibitem{szegedy2013intriguing}
Christian Szegedy, Wojciech Zaremba, Ilya Sutskever, Joan Bruna, Dumitru Erhan,
  Ian Goodfellow, and Rob Fergus.
\newblock Intriguing properties of neural networks.
\newblock In {\em International Conference on Learning Representations}, 2014.

\bibitem{akhtar2018defense}
Naveed Akhtar, Jian Liu, and Ajmal Mian.
\newblock Defense against universal adversarial perturbations.
\newblock In {\em Proceedings of the IEEE Conference on Computer Vision and
  Pattern Recognition}, pages 3389--3398, 2018.

\bibitem{Khrulkov_2018_CVPR}
Valentin Khrulkov and Ivan Oseledets.
\newblock Art of singular vectors and universal adversarial perturbations.
\newblock In {\em The IEEE Conference on Computer Vision and Pattern
  Recognition (CVPR)}, June 2018.

\bibitem{Akhtar_2018_CVPR}
Naveed Akhtar, Jian Liu, and Ajmal Mian.
\newblock Defense against universal adversarial perturbations.
\newblock In {\em The IEEE Conference on Computer Vision and Pattern
  Recognition (CVPR)}, June 2018.

\bibitem{xie2019feature}
Cihang Xie, Yuxin Wu, Laurens van~der Maaten, Alan~L Yuille, and Kaiming He.
\newblock Feature denoising for improving adversarial robustness.
\newblock In {\em Proceedings of the IEEE Conference on Computer Vision and
  Pattern Recognition}, pages 501--509, 2019.

\bibitem{Liao_2018_CVPR}
Fangzhou Liao, Ming Liang, Yinpeng Dong, Tianyu Pang, Xiaolin Hu, and Jun Zhu.
\newblock Defense against adversarial attacks using high-level representation
  guided denoiser.
\newblock In {\em The IEEE Conference on Computer Vision and Pattern
  Recognition (CVPR)}, June 2018.

\bibitem{grosse2017statistical}
Kathrin Grosse, Praveen Manoharan, Nicolas Papernot, Michael Backes, and
  Patrick McDaniel.
\newblock On the (statistical) detection of adversarial examples.
\newblock {\em arXiv preprint arXiv:1702.06280}, 2017.

\bibitem{hendrycks2016early}
Dan Hendrycks and Kevin Gimpel.
\newblock Early methods for detecting adversarial images.
\newblock {\em arXiv preprint arXiv:1608.00530}, 2016.

\bibitem{metzen2017detecting}
Jan~Hendrik Metzen, Tim Genewein, Volker Fischer, and Bastian Bischoff.
\newblock On detecting adversarial perturbations.
\newblock In {\em International Conference on Learning Representations}, 2017.

\bibitem{feinman2017detecting}
Reuben Feinman, Ryan~R Curtin, Saurabh Shintre, and Andrew~B Gardner.
\newblock Detecting adversarial samples from artifacts.
\newblock {\em arXiv preprint arXiv:1703.00410}, 2017.

\bibitem{meng2017magnet}
Dongyu Meng and Hao Chen.
\newblock Magnet: a two-pronged defense against adversarial examples.
\newblock In {\em Proceedings of the 2017 ACM SIGSAC Conference on Computer and
  Communications Security}, pages 135--147, 2017.

\bibitem{lusafetynet}
Jiajun Lu, Theerasit Issaranon, and David Forsyth.
\newblock Safetynet: Detecting and rejecting adversarial examples robustly.
\newblock In {\em Proceedings of the IEEE International Conference on Computer
  Vision}, pages 446--454, 2017.

\bibitem{liu2018fine}
Kang Liu, Brendan Dolan-Gavitt, and Siddharth Garg.
\newblock Fine-pruning: Defending against backdooring attacks on deep neural
  networks.
\newblock In {\em International Symposium on Research in Attacks, Intrusions,
  and Defenses}, pages 273--294. Springer, 2018.

\bibitem{chen2018detecting}
Bryant Chen, Wilka Carvalho, Nathalie Baracaldo, Heiko Ludwig, Benjamin
  Edwards, Taesung Lee, Ian Molloy, and Biplav Srivastava.
\newblock Detecting backdoor attacks on deep neural networks by activation
  clustering.
\newblock {\em arXiv preprint arXiv:1811.03728}, 2018.

\bibitem{chou2018sentinet}
Edward Chou, Florian Tram{\`e}r, Giancarlo Pellegrino, and Dan Boneh.
\newblock Sentinet: Detecting physical attacks against deep learning systems.
\newblock {\em arXiv preprint arXiv:1812.00292}, 2018.

\bibitem{wang2019neural}
Bolun Wang, Yuanshun Yao, Shawn Shan, Huiying Li, Bimal Viswanath, Haitao
  Zheng, and Ben~Y Zhao.
\newblock Neural cleanse: Identifying and mitigating backdoor attacks in neural
  networks.
\newblock In {\em 2019 IEEE Symposium on Security and Privacy (SP)}, pages
  707--723. IEEE, 2019.

\bibitem{chen2019deepinspect}
Huili Chen, Cheng Fu, Jishen Zhao, and Farinaz Koushanfar.
\newblock Deepinspect: A black-box trojan detection and mitigation framework
  for deep neural networks.
\newblock In {\em Proceedings of the 28th International Joint Conference on
  Artificial Intelligence. AAAI Press}, pages 4658--4664, 2019.

\bibitem{huang2019neuroninspect}
Xijie Huang, Moustafa Alzantot, and Mani Srivastava.
\newblock Neuroninspect: Detecting backdoors in neural networks via output
  explanations.
\newblock {\em arXiv preprint arXiv:1911.07399}, 2019.

\bibitem{guo2019tabor}
Wenbo Guo, Lun Wang, Xinyu Xing, Min Du, and Dawn Song.
\newblock Tabor: A highly accurate approach to inspecting and restoring trojan
  backdoors in ai systems.
\newblock {\em arXiv preprint arXiv:1908.01763}, 2019.

\bibitem{moosavi2016deepfool}
Seyed-Mohsen Moosavi-Dezfooli, Alhussein Fawzi, and Pascal Frossard.
\newblock Deepfool: a simple and accurate method to fool deep neural networks.
\newblock In {\em Proceedings of the IEEE conference on computer vision and
  pattern recognition}, pages 2574--2582, 2016.

\bibitem{howard2019searching}
Andrew Howard, Mark Sandler, Grace Chu, Liang-Chieh Chen, Bo~Chen, Mingxing
  Tan, Weijun Wang, Yukun Zhu, Ruoming Pang, Vijay Vasudevan, et~al.
\newblock Searching for mobilenetv3.
\newblock In {\em Proceedings of the IEEE International Conference on Computer
  Vision}, pages 1314--1324, 2019.

\bibitem{goodfellow2014explaining}
Ian~J. Goodfellow, Jonathon Shlens, and Christian Szegedy.
\newblock Explaining and harnessing adversarial examples.
\newblock In Yoshua Bengio and Yann LeCun, editors, {\em 3rd International
  Conference on Learning Representations, {ICLR} 2015, San Diego, CA, USA, May
  7-9, 2015, Conference Track Proceedings}, 2015.

\bibitem{hampel1974influence}
Frank~R Hampel.
\newblock The influence curve and its role in robust estimation.
\newblock {\em Journal of the american statistical association},
  69(346):383--393, 1974.

\end{thebibliography}

\section*{Supplementary Material}
\vspace{-3mm}
\subsection*{TrojAI Leaderboard Results on NIST-Round0 Dataset}
\vspace{-3mm}
\label{sec:leaderboard}

\begin{figure}[h!]
\centering
\includegraphics[width=0.9\columnwidth]{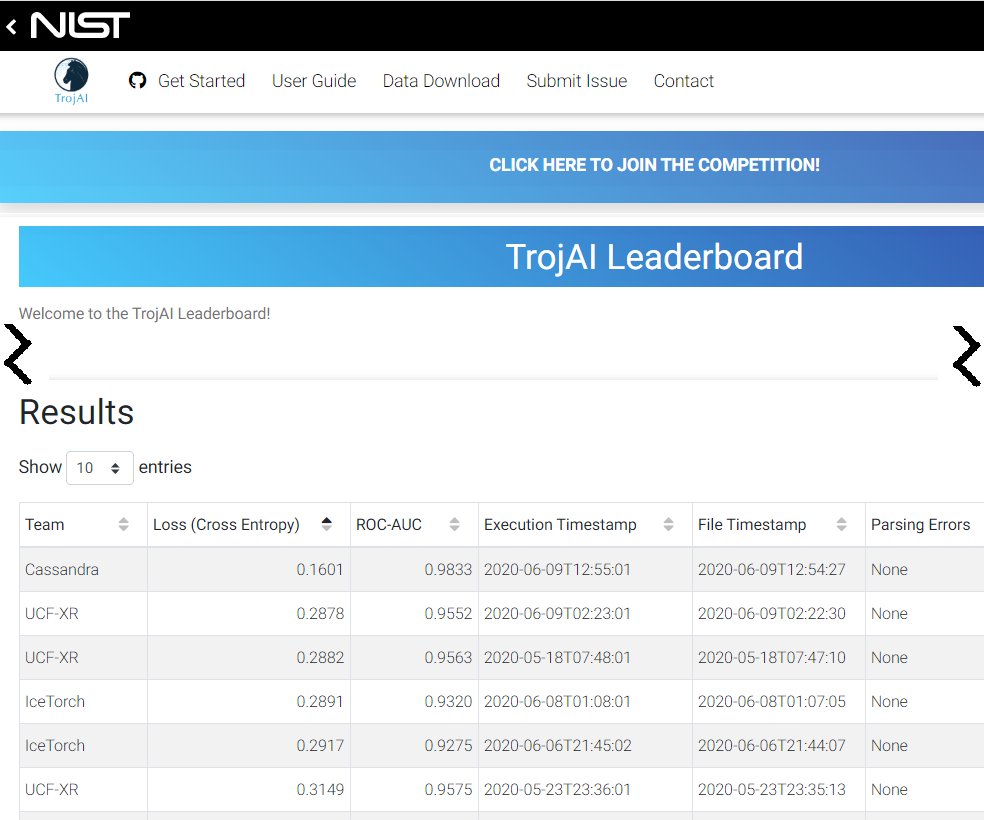}
%\vspace{-3mm}
\caption{A snapshot of the TrojAI Leaderboard shows our method on top.}
\label{fig:TrojAI_LB}
\end{figure}

Figure \ref{fig:TrojAI_LB} shows a snapshot of the TrojAI Leaderboard for NIST Round0 (see Section \ref{sec:nist_data} for dataset description). These results are compiled by the NIST server using a held-out test set not available publicly. The snapshot was taken at 21:30 hours on 10 June 2020 and adjusted to fit on this page without interfering with the results. We can see that Cassandra outperforms all other competitors by a significant margin. The best results in terms of Cross-Entropy Loss and ROC-AUC for each method are repeated in Table~\ref{tab:F1}.  Notice that we have the lowest loss and the highest ROC-AUC.

% The code will be released upon the acceptance of this paper
% \bibliographystyle{splncs}
%\bibliographystyle{unsrt}
%\bibliography{egbib}

\begin{table}[h!!!]
\caption{TrojAI Leaderboard results on NIST-Round0 dataset sorted by ROC (AUC). We have included the best results in terms of Loss and ROC for our competitors. Best results in each column are in bold. Note that we cannot access the F1 score, Precision and Recall for other methods.}
\begin{center}
\label{tab:F1}
\begin{tabular}{|l|c|c|c|c|c|}
\hline
\textbf{Team} & \textbf{Loss (Cross-Entropy)} & \textbf{ROC (AUC)} &\textbf{F1} &\textbf{Precision} & \textbf{Recall} \\ \hline\hline
% &MNIST &  92.5  & 88.5 & 90.5 &&\\ \hline
 Cassandra   & {\bf 0.1601} & {\bf 0.9833} & {\bf0.9746}&{\bf0.9897} &{\bf0.9600}\\ \hline
UCF-XR     & 0.3149 & 0.9575 &-&-&-  \\ \hline
UCF-XR     & 0.2882 & 0.9563 &-&-&- \\ \hline
IceTorch   & 0.2891 & 0.9320 &-&-&-\\ \hline
IceTorch   & 0.2917 & 0.9275 &-&-&-\\ \hline
\end{tabular}
\end{center}
\end{table}

%\section{Supplementary material}

\subsection*{ MNIST Model Generation}

\subsubsection*{Clean Model Generation}

 The data is split into training set : 60,000 images (6,000 images per class), and test set: 10,000 images (1,000 from each class). The clean data are used for training 300 benign/clean models with three architecture types (ModdedBadnet, Badnet and ModdedLenet5net), each with 100 models (see Table~\ref{table:model-table1}) .

\subsubsection*{Trojaned Model Generation} 

\textbf{Clean Data}: The MNIST dataset has 10 classes with  70,000 clean images (without triggers). 

\textbf{Triggered Data}: Two types of triggers, Type I and Type II (see Figure in main paper) were inserted into  images of MNIST dataset.
The Triggered MNIST data was combined with clean data to generate Trojaned models. The following three data splits were used in our experiments:\\
Data split 1: training: 60,000 (triggered data: 10\%),  testing:10,000 (triggered data: 10\%).\\
Data split 2: training: 60,000 (triggered data: 15\%),  testing:10,000 (triggered data: 15\%).\\
Data split 3: training: 60,000 (triggered data: 20\%),  testing:10,000 (triggered data: 20\%).
    
\textbf{Models}: 
In addition to the 300 benign models, another 600 Trojaned models of the same three architectures (ModdedBadnet, Badnet and ModdedLenet5net) were generated. 
 Trojaned models were trained by the Triggered MNIST data and clean data where the proportion of triggered data varied as 10\%, 15\% and 20\%. Table~\ref{table:model-table1} shows the details of both clean  and infected models trained for any-to-any Trojan attack. Any-to-one attack models were generated similar to any-to-any models.
300 Trojaned models were trained by any-to-any targeted attack, and another 300 were trained for any-to-one targeted attack.

Evaluations of ModdedBadnet, Badnet and ModdedLeNet5 models are shown in Table~\ref{tab:MNIST_model-A} for any-to-any attack and in Table~\ref{tab:MNIST_model-B} for any-to-one targeted attack. The clean models and Trojaned models both have high classification accuracy when the test data is clean. The clean  models also have high classification accuracy when the test data is triggered. Since there is no Trojan in the clean model, the triggered image samples are correctly classified. However, for the Trojaned models, the classification accuracy (100 $-$ Attack Success Rate) for triggered data is low since the triggered images are misclassified. The tables show Attack Success Rates only for the triggered data which is very high. These results imply that the Trojan (backdoor) was successfully inserted into the models.

\begin{table}[htbp]
	\centering
	\caption{Three models trained on Triggered MNIST dataset. Half the models are for any-to-any attack and half are for any-to-one attack. For the latter case each model only has one target class.}
	\footnotesize
%	\begin{adjustbox}{width=0.9\linewidth,height=20 \lineheight, center}
	
		\begin{tabular}{|l|l|c|c|c|c|}
 		\hline
%    Input Samples & Training Set models & num of Training set models & Testing Set models & num of test set models & Classification Accuracy. \\
  {\bf  Model name} & {\bf Model Architecture} & {\bf Trigger} & {\bf Triggered data } & {\bf \#} \\
		\hline
		\hline
   ModdedBadNet & 2 Conv + 1 Dense & Type I, II & 10\%, 15\% and 20\% & 100+100\\
		\hline
   BadNet       & 2 Conv + 2 Dense & Type I, II & 10\%, 15\% and 20\% & 100+100\\
		\hline
   ModdedLeNet5 & 3 Conv + 2 Dense & Type I, II & 10\%, 15\% and 20\% & 100+100 \\
% 		\hline
%   Resnet & Resnet18 & Type I, II & 10\%, 15\% and 20\%  & 75\\
	
 		\hline
		\end{tabular}
%	\end{adjustbox}
	\label{table:model-table1}
% \end{table*}

\vspace{5mm}

%  \begin{table}[h!]
        \centering
        \caption{Attack success rate and classification accuracy for three types of trojaned models (any-to-any attack) for Triggered MNIST dataset. Success rate is the proportion of images for which predictions by the Trojaned model is changed to an incorrect label.}
        \begin{tabular}{|l|c|c|c|c|c|}
        \hline
        %\abovespace\belowspace
         &  \multicolumn{4}{c|}{\textbf{Trojaned Model}} & \textbf{Clean Model} \\
        \hline

         & \multicolumn{2}{c|}{\textbf{Attack Success Rate}} & \multicolumn{3}{c|}{\textbf{Classification Accuracy}}  \\
        \hline
        \textbf{Model Type}  & \textbf{Trigger I} & \textbf{Trigger II}  & \textbf{Trigger I} & \textbf{Trigger II} & \\
        \hline\hline
        BadNet          & 98.7 & 98.7 & 98.9 & 99.0 & 99.1 \\
        ModdedBadNet    & 97.3 & 97.6 & 97.2 & 96.5 & 98.8 \\
        ModdedLeNet5net & 97.8 & 98.6 & 98.0 & 97.3 & 98.7 \\
        % Resnet18 & 98.5 &98.8  & 99.1 &98.9 & 99.2 \\
        \hline
    \end{tabular}
        \label{tab:MNIST_model-A}%
\end{table}

% \vspace{5mm}

 \begin{table}[h!]
        \centering
        \caption{Attack success rate and classification accuracy for three types of trojaned models (any-to-one targeted attack) on the Triggered MNIST dataset. Success rate is the proportion of images that changed label to the target class for Trojaned model.}
        \begin{tabular}{|l|c|c|c|c|}
        \hline
        %\abovespace\belowspace
         &  \multicolumn{4}{c|}{\textbf{Trojaned Model}}  \\
        \hline

         & \multicolumn{2}{c|}{\textbf{Attack Success Rate}} & \multicolumn{2}{c|}{\textbf{Classification Accuracy}} \\
        \hline
         \textbf{Model Type} & \textbf{Trigger I} & \textbf{Trigger II}  & \textbf{Trigger I} & \textbf{Trigger II} \\
        \hline\hline
        Badnet  &99.1 & 99.0  & 98.8 & 98.9 \\
        ModdedBadnet &98.5 & 98.3 & 97.6 & 97.4\\
        ModdedLenet5net &98.8 & 98.4 & 98.0 & 97.5 \\
        % Resnet18 & 98.5 &98.8  & 99.1 &98.9 & 99.2 \\
        \hline
    \end{tabular}
        \label{tab:MNIST_model-B}%
\end{table}

% \begin{figure}[t]
% \centering
% \includegraphics[width=0.6\columnwidth]{figs/mbms4.png}
% \vspace{-5mm}
% \caption{Universal Perturbations for a infected ResNet model. (a) is the  universal perturbation generated by using the full dataset (400 images). (b) shows universal perturbations generated by splitting the data into 10 batches (40 images each). The feature distribution of the multi-batch universal perturbations reveal the characteristic of perturbations from Trojaned models.}
% \label{fig:mbms}
% \end{figure}

% \begin{figure}[t]
% \centering
% \includegraphics[width=\linewidth]{figs/result2.png}
% \vspace{-5mm}
% \caption{TrojAI Leaderboard result.}
% \label{fig:mbms}
% \end{figure}

\hfill

\newpage

\subsubsection*{NIST Round0 and NIST Round1 Datasets}
\label{sec:nist_data}
The NIST datasets consist of CNN classification models for traffic sign signals. Half of the models are benign models and half are Trojaned models. The models have three architectures namely, Inception-v3, DenseNet-121, and ResNet50. The models were trained on synthetically created image data of artificial traffic signs superimposed on road background scenes. The Trojaned models have been poisoned with triggers of different color, size and shape. \textbf{Round0} dataset consists of 200 models, while \textbf{Round1} dataset has 1,000 models. NIST also holds a sequestered test dataset to evaluate models. For that, models must be uploaded to the TrojAI Leaderboard website. Section \ref{sec:leaderboard} and Table \ref{tab:F1} discuss our results on the TrojAI Leaderboard.

% Note that NIST does not provide any information on the type or size of these triggers.

%\textbf{NIST-Round0} and \textbf{NIST-Round1} datasets are both from the same distribution.

Table \ref{tab:NIST0_model} and Table \ref{tab:NIST1_model} show the model details and the performance of the three architecture types present in the NIST Round0 and Round1 datasets.  Notice that the Trojan infected models have accuracy at par with the clean models and yet they have a very high attack success rate on the triggered data.

\begin{table}[h]
        \centering
        \caption{Attack success rate (for Trojan trigger infused data) and top-1 classification accuracy (for clean data) for %three types of trojan infected and clean models from the
        NIST-Round0 dataset. Success rate is the proportion of images for which the prediction changes to the target label in Trojaned models.}
        \begin{tabular}{|l|c|c|c|c|}
        \hline
        %\abovespace\belowspace
         & \multicolumn{2}{c|}{\textbf{Trojan Infected Model}} & \textbf{Clean Model} &  \\
        \hline
        \textbf{Model Type} & \textbf{Attack Success Rate} & \multicolumn{2}{c|}{\textbf{Classification Accuracy}} & \# \textbf{models} \\
        \hline \hline
        DenseNet-121 & 99.82 & 99.76 & 99.90 & 63 \\
        Inception-v3 & 99.87 & 99.69 & 99.75 & 69\\
        ResNet50    & 99.80 & 99.68 & 99.76 & 68 \\
        \hline
        \# \textbf{models} & \multicolumn{2}{c|}{100} & 100 & \textbf{200} \\
        % Resnet18 & 98.5 &98.8  & 99.1 &98.9 & 99.2 \\
        \hline
    \end{tabular}
        \label{tab:NIST0_model}%
\end{table}

\vspace{5mm}

\begin{table}[h]
%\begin{table}[h]
        \centering
        \caption{Attack success rate (for Trojan trigger infused data) and top-1 classification accuracy (for clean data) for three types of Trojaned and clean models from the NIST-Round1 dataset. Success rate is the proportion of images for which the prediction changes to the target label in Trojaned models.}
        \begin{tabular}{|l|c|c|c|c|}
        \hline
        %\abovespace\belowspace
        &  \multicolumn{2}{c|}{\textbf{Trojan Infected Model}} & \textbf{Clean Model} & \\
        \hline
        \textbf{Model Type}  & \textbf{Attack Success Rate} & \multicolumn{2}{c|}{\textbf{Classification Accuracy}} & \# \textbf{models}  \\
        \hline \hline
        DenseNet-121  & 99.88 & 99.81 & 99.88 & 313 \\
        Inception-v3  & 99.84 & 99.85 & 99.89 & 250 \\
        ResNet50     & 99.58 & 99.81 & 99.83 & 437 \\
        % Resnet18 & 98.5 &98.8  & 99.1 &98.9 & 99.2 \\
        \hline
        \# \textbf{models} & \multicolumn{2}{c|}{500} & 500 & \textbf{1000} \\
        \hline
    \end{tabular}
        \label{tab:NIST1_model}%
\end{table}

\subsection*{Target Class Detection Algorithm}
The procedure for target class prediction is given in Algorithm \ref{alg:algorithm1}.

\begin{algorithm}[H]
    \KwData{Query model}
    \KwResult{$P(Trojan)$ and Target Class}
    % initialization; \\
    Stage One: Use Trojan Detection network to get $P(Trojan)$; \\
    \eIf{$P(Trojan)$ >= 0.5}
    {\For{$C_i\gets0$ \KwTo $C$ }
        {
            use FGSM to calculate adversarial perturbations with $C_i$ as the target class;\\
            compute attack difficulty ($\sigma_i = \frac{L_{1}{Norm}}{Fooling Rate}$) for perturbation;\\
        }
        $TargetClass$ = perform outlier detection over the attack difficulties $\sigma_i$s;\\
        output $P(Trojan)$ and target class prediction;\\
    }
    {
        output $P(Trojan)$ and target class(None);\\
    }
 
 \caption{Two-stage method to detect a Trojan infected model and predict its target class using only clean image samples.}
 \label{alg:algorithm1}
\end{algorithm}

% \section*{Acknowledgments}

% Use unnumbered third level headings for the acknowledgments. All acknowledgments
% go at the end of the paper. Do not include acknowledgments in the anonymized
% submission, only in the final paper.

% \cleardoublepage
% \section*{References}

% References follow the acknowledgments. Use unnumbered first-level heading for
% the references. Any choice of citation style is acceptable as long as you are
% consistent. It is permissible to reduce the font size to \verb+small+ (9 point)
% when listing the references. {\bf Remember that you can use more than eight
%   pages as long as the additional pages contain \emph{only} cited references.}
% \medskip

% \small

% [1] Alexander, J.A.\ \& Mozer, M.C.\ (1995) Template-based algorithms for
% connectioNIST rule extraction. In G.\ Tesauro, D.S.\ Touretzky and T.K.\ Leen
% (eds.), {\it Advances in neural Information Processing Systems 7},
% pp.\ 609--616. Cambridge, MA: MIT Press.

% [2] Bower, J.M.\ \& Beeman, D.\ (1995) {\it The Book of GEnESIS: Exploring
%   Realistic neural Models with the GEneral nEural SImulation System.}  new York:
% TELOS/Springer--Verlag.

% [3] Hasselmo, M.E., Schnell, E.\ \& Barkai, E.\ (1995) Dynamics of learning and
% recall at excitatory recurrent synapses and cholinergic modulation in rat
% hippocampal region CA3. {\it Journal of neuroscience} {\bf 15}(7):5249-5262.

\end{document}